\documentclass[conference]{IEEEtran}
\IEEEoverridecommandlockouts

\usepackage{amsmath,graphicx}
\usepackage{amssymb}
\usepackage{booktabs}
\usepackage{acronym, color, soul}
\usepackage{multirow, bm}
\usepackage{adjustbox}

\usepackage{amsfonts}
\usepackage{algorithmic}
\usepackage{algorithm}
\usepackage{array}
\usepackage[caption=false,font=normalsize,labelfont=sf,textfont=sf]{subfig}
\usepackage{textcomp}
\usepackage{stfloats}
\usepackage{url}
\usepackage{verbatim}
\usepackage{cite}
\usepackage{hyperref}
\usepackage{booktabs}
\usepackage{xspace}
\usepackage{xcolor}

\def\BibTeX{{\rm B\kern-.05em{\sc i\kern-.025em b}\kern-.08em
    T\kern-.1667em\lower.7ex\hbox{E}\kern-.125emX}}

\makeatletter
\DeclareRobustCommand\onedot{\futurelet\@let@token\@onedot}
\def\@onedot{\ifx\@let@token.\else.\null\fi\xspace}
\def\etal{\emph{et al}\onedot}

\newacro{convnet}[ConvNet]{convolutional neural network}
\newacro{dnn}[DNN]{deep neural network}
\newacro{ann}[ANN]{artificial neural network}
\newacro{rnn}[RNN]{recurrent neural network}
\newacro{vae}[VAE]{variational autoencoder}

\newacro{vvc}[VVC]{versatile video coding}
\newacro{hevc}[HEVC]{high-efficiency video coding}
\newacro{bpg}[BPG]{better portable graphics}

\newacro{gdn}[GDN]{generalized divisive normalization}
\newacro{nlp}[NLP]{natural language processing}
\newacro{mha}[MHA]{multi-headed attention}
\newacro{gmm}[GMM]{gaussian mixture model}
\newacro{charm}[ChARM]{channel-wise auto-regressive model}

\newacro{tfc}[TFC]{tensorFlow compression}

\title{ConvNeXt-ChARM: ConvNeXt-based Transform for Efficient Neural Image Compression}

\author{
Ahmed Ghorbel \qquad Wassim Hamidouche \qquad Luce Morin 
\thanks{Ahmed Ghorbel and Luce Morin were with Univ. Rennes, INSA Rennes, CNRS, IETR - UMR 6164, Rennes, France, e-mail: (Ahmed.Ghorbel; Luce.Morin)@insa-rennes.fr.}
\thanks{Wassim Hamidouche was with Technology Innovation Institute, Masdar City, P.O Box 9639, Abu Dhabi, UAE, e-mail: Wassim.Hamidouche@tii.ae.}
\thanks{This work has been supported by Région Bretagne and Rennes Ville et Métropole under the DEEPTEC project.}
}

\begin{document}
\maketitle
%
\begin{abstract}
Over the last few years, neural image compression has gained wide attention from research and industry, yielding promising end-to-end deep neural codecs outperforming their conventional counterparts in rate-distortion performance.
Despite significant advancement, current methods, including attention-based transform coding, still need to be improved in reducing the coding rate while preserving the reconstruction fidelity, especially in non-homogeneous textured image areas. Those models also require more parameters and a higher decoding time.
To tackle the above challenges, we propose ConvNeXt-ChARM, an efficient ConvNeXt-based transform coding framework, paired with a compute-efficient channel-wise auto-regressive prior to capturing both global and local contexts from the hyper and quantized latent representations. The proposed architecture can be optimized end-to-end to fully exploit the context information and extract compact latent representation while reconstructing higher-quality images.
Experimental results on four widely-used datasets showed that ConvNeXt-ChARM brings consistent and significant BD-rate (PSNR) reductions estimated on average to $5.24\%$ and $1.22\%$ over the \ac{vvc} reference encoder (VTM-18.0) and the state-of-the-art learned image compression method SwinT-ChARM, respectively.
Moreover, we provide model scaling studies to verify the computational efficiency of our approach and conduct several objective and subjective analyses to bring to the fore the performance gap between the next generation ConvNet, namely ConvNeXt, and Swin Transformer.
All materials, including the source code of SwinT-ChARM, will be made publicly accessible upon acceptance for reproducible research. 
\end{abstract}

\section{Introduction}
\label{intro}
Visual information is crucial in human development, communication, and engagement, and its compression is necessary for effective storage and transmission over constrained wireless/wireline channels.
\begin{figure}[t]
\centering
\includegraphics[width=0.49\textwidth]{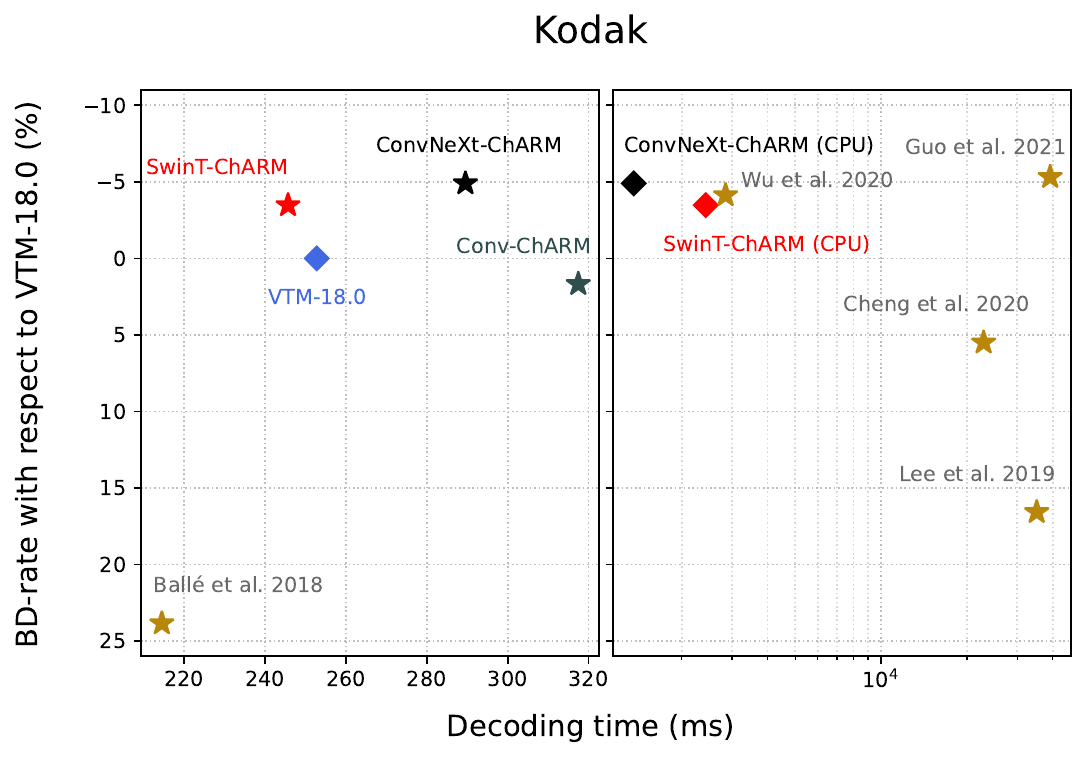}
\caption{BD-rate (\%) versus decoding time (ms) on the Kodak dataset. Left-top is better. Star and diamond markers refer to decoding on GPU and CPU, respectively.}
\vspace{-5mm}
\label{bdr_vs_dt}
\end{figure}
Thus, thinking about new lossy image compression approaches is a goldmine for scientific research. The goal is to reduce an image file size by permanently removing less critical information, particularly redundant data and high frequencies, to obtain the most compact bit-stream representation while preserving a certain level of visual fidelity. Nevertheless, the high compress rate and low distortion are fundamentally opposing objectives involving optimizing the rate-distortion tradeoff.

Conventional image and video compression standards including JPEG~\cite{monteiro2016light}, JPEG2000~\cite{rabbani2002overview}, H.265/\ac{hevc}~\cite{sullivan2012overview}, and H.266/\ac{vvc}~\cite{sullivan2020versatile}, rely on hand-crafted creativity to present module-based encoder/decoder block diagram. In addition, these codecs employ intra-prediction, fixed transform matrices, quantization, context-adaptive arithmetic coders, and various in-loop filters to reduce spatial and statistical redundancies, and alleviate coding artifacts. However, it has taken several years to standardize a conventional codec. Moreover, existing image compression standards are not anticipated to be an ideal and global solution for all types of image content due to the rapid development of new image formats and the growth of high-resolution mobile devices. 

Lossy image compression consists of three modular parts: transform, quantization, and entropy coding. Each of these components can be represented as follows: i) autoencoders as flexible nonlinear transforms where the encoder (i.e., analysis transform) extracts latent representation from an input image and the decoder (i.e., synthesis transform) reconstructs the image from the decoded latent, ii) various differentiable quantization approaches which encode the latent into bitstream through arithmetic coding algorithms, iii) deep generative models as potent learnable entropy models estimating the conditional probability distribution of the latent to reduce the rate. Moreover, these three components can be optimized with end-to-end training by reducing the joint loss of the distortion between the original image and its reconstruction and the rate needed to transmit the bitstream of latent representation.

Thanks to recent advances in deep learning, we have seen many works exploring the potential of \acp{ann} to form various learned image and video compression frameworks. Over the past two years, the performance of neural compression has steadily improved thanks to the prior line of study, reaching or outperforming state-of-the-art conventional codecs. 
Some previous works use local context \cite{minnen2018joint,lee2019extended,mentzer2018conditional}, or additional side information \cite{balle2018variational,hu2020coarse,minnen2018image} to capture short-range spatial dependencies, and others use non-local mechanism \cite{cheng2020learned,li2021learning,qian2020learning,chen2021end} as long-range spatial dependencies. Recently, Toderici \etal \cite{toderici2022high} proposed a generative compression method achieving high-quality reconstructions, Minnen \etal \cite{minnen2020channel} introduced channel-conditioning and latent residual prediction taking advantage of an entropy-constrained model that uses both forward and backward adaptations, and Zhu \etal \cite{zhu2021transformer} replaced all convolutions in the \ac{charm} prior approach \cite{minnen2020channel} with Swin Transformer \cite{liu2021swin} blocks, Zou \etal  \cite{zou2022devil} combined the local-aware attention mechanism with the global-related feature learning and proposed a window-based attention module, Koyuncu et al. \cite{koyuncu2022contextformer} proposed a Transformer-based context model, which generalizes the standard attention mechanism to spatio-channel attention, Zhu \etal \cite{zhu2022unified} proposed a probabilistic vector quantization with cascaded estimation under a multi-codebooks structure, Kim \etal \cite{kim2022joint} exploited the joint global and local hyperpriors information in a content-dependent manner using an attention mechanism, and He \etal \cite{he2022elic} adopted stacked residual blocks as nonlinear transform and multi-dimension entropy estimation model.

One of the main challenges of learned transform coding is the ability to identify the crucial information necessary for the reconstruction, knowing that information overlooked during encoding is usually lost and unrecoverable for decoding. Another main challenge is the tradeoff between performance and decoding speed. While the existing approaches improve the transform and entropy coding accuracy, they remain limited by the higher decoding runtime and excessive model complexity leading to an ineffective real-world use. Finally, we found that attention-based networks taking advantage of attention mechanisms to capture global dependencies, such as Swin Transformer \cite{liu2021swin}, have over-smoothed and contain undesirable artifacts at low bitrates. Furthermore, the global semantic information in image compression is less effective than in other computer vision tasks \cite{zou2022devil}.

In this paper, we propose a nonlinear transform built on ConvNeXt blocks with additional down and up sampling layers and paired with a \ac{charm} prior, namely ConvNeXt-ChARM. Recently proposed in \cite{liu2022convnet}, ConvNeXt is defined as a modernized ResNet architecture toward the design of a vision Transformer, which competes favorably with Transformers in terms of efficiency, achieving state-of-the-art on ImageNet classification task \cite{deng2009imagenet} and outperforming Swin Transformer on COCO detection \cite{lin2014microsoft} and ADE20K segmentation \cite{zhou2017scene} challenges while maintaining the maturity and simplicity of \acp{convnet} \cite{liu2022convnet}. The contributions of this paper are summarized as follows:
\begin{itemize}
\item We propose a learned image compression model that leverages a stack of ConvNeXt blocks with down and up-sampling layers for extracting contextualized and nonlinear information for effective latent decorrelation. We maintain the convolution strengths like sliding window strategy for computations sharing, translation equivariance as a built-in inductive bias, and the local nature of features, which are intrinsic to providing a better spatial representation.
\item We apply ConvNeXt-based transform coding layers for generating and decoding both latent and hyper-latent to consciously and subtly balance the importance of feature compression through the end-to-end learning framework. 
\item We conduct experiments on four widely-used evaluation datasets to explore possible coding gain sources and demonstrate the effectiveness of ConvNeXt-ChARM. In addition, we carried out a model scaling analysis to compare the complexity of ConvNeXt and Swin Transformer.
\end{itemize}
Extensive experiments validate that the proposed ConvNeXt-ChARM achieves state-of-the-art compression performance, as illustrated in Figure~\ref{bdr_vs_dt}, outperforming conventional and learned image compression methods in the tradeoff between coding efficiency and decoder complexity.

The rest of this paper is organized as follows. Section~\ref{methd} presents our overall framework along with a detailed description of the proposed architecture. Next, we dedicate Section~\ref{result} to describe and analyze the experimental results. Finally, Section~\ref{concl} concludes the paper.

\begin{figure*}[htbp!]
  \centering
  \includegraphics[width=1\linewidth]{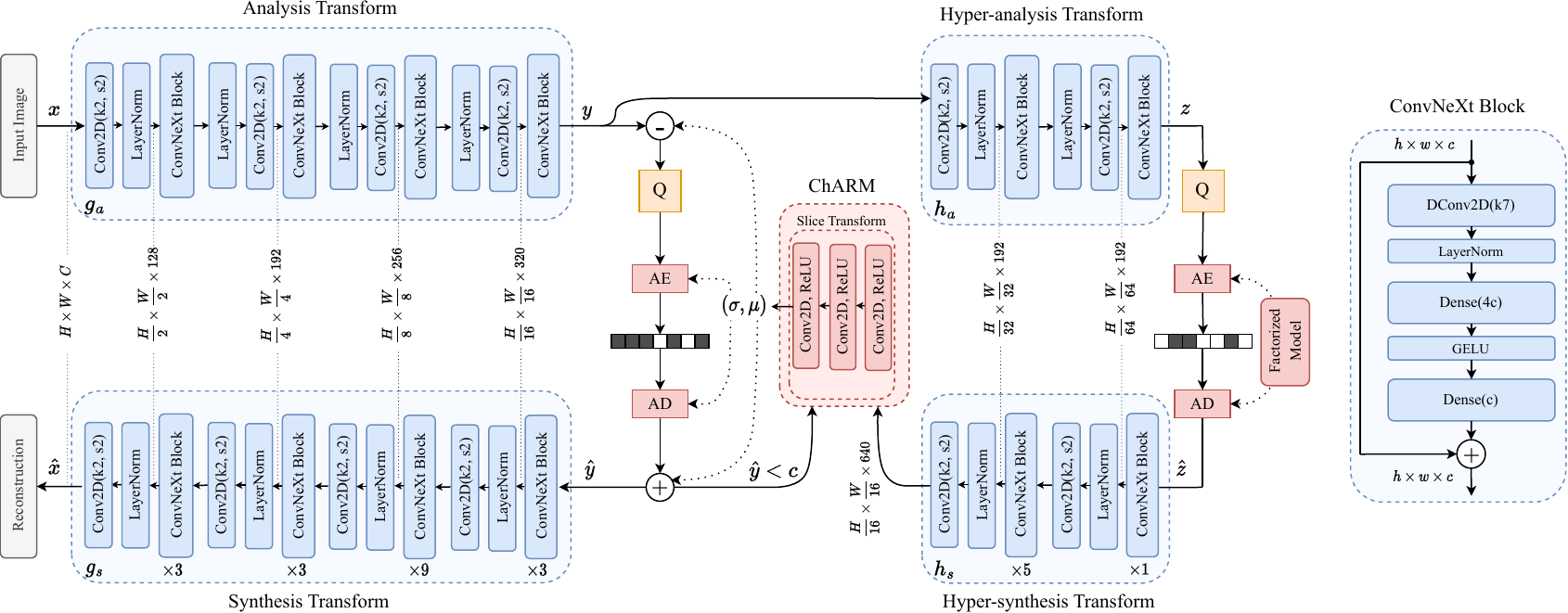}
  \caption{Overall ConvNeXt-ChARM Framework. We illustrate the image compression diagram of our ConvNeXt-ChARM with hyperprior and channel-wise auto-regressive context model. We also present the ConvNeXt block used in both transform and hyper-transform coding for an end-to-end feature aggregation.}
  \label{fig:convnextCharm}
\end{figure*}
\section{Proposed ConvNeXt-ChARM Model}
\label{methd}
\subsection{Problem Formulation}
The objective of learned image compression is to minimize the distortion between the original image and its reconstruction under a specific distortion-controlling hyper-parameter. Assuming an input image $\bm{x}$, the analysis transform $g_{a}$, with parameter $\phi_{g}$, removes the image spatial redundancies and generates the latent representation $\bm{y}$. Then, this latent is quantized to the discrete code $\hat{\bm{y}}$ using the quantization operator $\lceil$.$\rfloor$, from which a synthesis transform $g_{s}$, with parameter $\theta_{g}$, reconstructs the image denoted by $\hat{\bm{x}}$. The overall process can be formulated as follows:
\begin{equation}
\begin{aligned}
\bm{y} &= g_{a}( \bm{x} \mid \phi_{g}), \\
\hat{\bm{y}} &= \lceil \bm{y} \rfloor, \\
\hat{\bm{x}} &= g_{s}(\hat{\bm{y}} \mid \theta_{g}).
\end{aligned}
\end{equation}

A hyperprior model composed of a hyper-analysis and hyper-synthesis transforms $(h_{a}, h_{s})$ with parameters $(\phi_{h}, \theta_{h})$ is usually used to reduce the statistical redundancy among latent variables. In particular, this hyperprior model assigns a few extra bits as side information to transmit some spatial structure information and helps to learn an accurate entropy model. The hyperprior generation can be summarized as follows:
\begin{equation}
\begin{aligned}
\bm{z} &= h_{a}(\bm{y} \mid \phi_{h}), \\
\hat{\bm{z}} &= \lceil \bm{z} \rfloor, \\
p_{\hat{\bm{y}} \mid \hat{\bm{z}}}(\hat{\bm{y}} \mid \hat{\bm{z}}) & \leftarrow h_{s}(\hat{\bm{z}} \mid \theta_{h}).
\end{aligned}
\end{equation}
Transform and quantization introduce a distortion $D = MSE(\bm{x}, \hat{\bm{x}})$, for mean squared error (MSE) optimization that measures the reconstruction quality with an estimated bitrate $R$, corresponding to the expected rate of the quantized latents and hyper-latents, as described bellow: 
\begin{equation}
R =  \mathbb{E} \left [-\log _{2}(p_{\hat{\bm{y}} \mid \hat{\bm{z}}}(\hat{\bm{y}} \mid \hat{\bm{z}})) -\log _2(p_{\hat{\bm{z}}}(\hat{\bm{z}}))\right ].
\end{equation}

Representing $(g_{a},g_{s})$, $(h_{a},h_{s})$, and entropy model by \acp{dnn} enables jointly optimizing the end-to-end model by minimizing the rate-distortion tradeoff $\mathcal{L}$, giving a rate-controlling hyper-parameter $\lambda$. This optimization problem can be presented as follows: 
\begin{equation}
\begin{split}
\mathcal{L} & = R+\lambda D, \\
& = \underbrace{ \mathbb{H}(\bm{\hat{y}}) + \mathbb{H} (\bm{\hat{z}}) }_R + \lambda \, MSE(\bm{x}, \bm{\hat{x}}),
\end{split}
\end{equation}
where $\mathbb{H}$ stands for the entropy.
\subsection{ConvNeXt-ChARM network architecture}
To better parameterize the distributions of the quantized latent features with a more accurate and flexible entropy model, we adopted the \ac{charm} prior approach proposed in \cite{minnen2020channel} to build an efficient ConvNeXt-based learning image compression model with strong compression performance. As shown in Figure~\ref{fig:convnextCharm}, the analysis/synthesis transform $(g_{a},g_{s})$ of our design consists of a combination of down and up-sampling blocks and ConvNeXt encoding/decoding blocks~\cite{liu2022convnet}, respectively. Down and up-sampling blocks are performed using Conv2D and Normalisation layers sequentially. The architectures for hyper-transforms $(h_{a},h_{s})$ are similar to $(g_{a},g_{s})$ with different stages and configurations.

\subsection{ConvNeXt design description}
Globally, ConvNeXt incorporates a series of architectural choices from a Swin Transformer while maintaining the network's simplicity as a standard \ac{convnet} without introducing any attention-based modules. These design decisions can be summarized as follows: macro design, ResNeXt's grouped convolution, inverted bottleneck, large kernel size, and various layer-wise micro designs. In Figure~\ref{fig:convnextCharm}, we illustrates the ConvNeXt block, where the DConv2D(.) refers for the a depthwise 2D convolution, LayerNorm for the layer normalization, Dense(.) for the densely-connected NN layer, and GELU for the activation function.
\\\\
\textbf{Macro design}:
The stage compute ratio is adjusted from (3, 4, 6, 3) in ResNet-50 to (3, 3, 9, 3), which also aligns the FLOPs with Swin-T. In addition, the ResNet-style stem cell is replaced with a patchify layer implemented using a 2$\times$2, stride two non-overlapping convolutional layers with an additional normalization layer to help stabilize the training. In ConvNeXt-ChARM diagram, we adopted the (3, 3, 9, 3) and (5, 1) as stage compute ratios for transforms and hyper-transforms, respectively. 
\\\\
\textbf{Depthwise convolution}:
The ConvNeXt block uses a depthwise convolution, a special case of grouped convolution used in ResNeXt \cite{xie2017aggregated}, where the number of groups is equal to the considered channels. This is similar to the weighted sum operation in self-attention, which operates by mixing information only in the spatial dimension.
\\\\
\textbf{Inverted bottleneck}:
Similar to Transformers, ConvNeXt is designed with an inverted bottleneck block, where the hidden dimension of the residual block is four times wider than the input dimension. As illustrated in the ConvNeXt block Figure~\ref{fig:convnextCharm}, the first dense layer is 4 times wider then the second one.
\\\\
\textbf{large kernel}:
One of the most distinguishing aspects of Swin Transformers is their local window in the self-attention block. The information is propagated across windows, which enables each layer to have a global receptive field. The local window is at least 7$\times$7 sized, which is still more extensive than the 3$\times$3 ResNeXt kernel size. Therefore, ConvNeXt adopted large kernel-sized convolutions by using a 7$\times$7 depthwise 2D convolution layer in each block. This allows our ConvNeXt-ChARM model to capture global contexts in both latents and hyper-latents, which are intrinsic to providing a better spatial representation.
\\\\
\textbf{Micro design}:
In ConvNeXt's micro-design, several per-layer enhancements are applied in each block, by using: a single Gaussian error linear unit (GELU) activation function (instead of numerous ReLU), using a single LayerNorm as normalization choice (instead of numerous BatchNorm), and using separate down-sampling layers between stages.

\section{Results}
\label{result}
First, we briefly describe used datasets with the implementation details. Then, we assess the compression efficiency of our method with a rate-distortion comparison and compute the average bitrate savings on four commonly-used evaluation datasets. We further elaborate a model scaling and complexity study to consistently examine the effectiveness of our proposed method against pioneering ones.

\subsection{Experimental Setup}
{\bf Datasets.}
The training set of the CLIC2020 dataset is used to train the proposed ConvNeXt-ChARM model. This dataset contains a mix of professional and user-generated content images in RGB color and grayscale formats. We evaluate image compression models on four datasets, including Kodak~\cite{testsets}, Tecnick~\cite{testsets}, JPEG-AI~\cite{testsets}, and the testing set of CLIC{21}~\cite{testsets}. For a fair comparison, all images are cropped to the highest possible multiples of 256 to avoid padding for neural codecs.
\\\\
{\bf Implementation details.}
We implemented all models in TensorFlow using \ac{tfc} library \cite{tfc_github}, and the experimental study was carried out on an RTX 5000 Ti GPU. All models were trained on the same CLIC2020 training set with 3.5M steps using the ADAM optimizer with parameters $\beta_1=0.9$ and $\beta_2=0.999$. The initial learning rate is set to $10^{-4}$ and drops to $10^{-5}$ for another 100k iterations, and $L=R+\lambda{D}$ as loss function.
The MSE is used as the distortion metric in RGB color space. Each batch contains eight random 256 $\times$ 256 crops from training images. To cover a wide range of rate and distortion, for our proposed method, we trained five models with $\lambda \in \{0.006, 0.009, 0.020, 0.050, 0.150\}$. Regarding the evaluation on CPU, we used an Intel(R) Xeon(R) W-2145 @ 3.70GHz.   
\\\\
{\bf Baselines.\footnote{For a fair comparison, we only considered SwinT-ChARM \cite{zhu2021transformer} from the state-of-the-art models \cite{zhu2021transformer,koyuncu2022contextformer,zou2022devil,zhu2022unified,kim2022joint,he2022elic}, due to the technical feasibility of models training and evaluation under the same conditions and in an adequate time.}}
We compare our approach with the state-of-art neural compression method SwinT-ChARM proposed by Zhu \etal \cite{zhu2021transformer}, and non-neural compression methods, including \ac{bpg}(4:4:4), and the most up-to-date \ac{vvc} official Test Model VTM-18.0 in All-Intra profile configuration.

\subsection{Rate-Distortion coding performance}
\begin{figure}[t]
\centering
\includegraphics[width=0.45\textwidth]{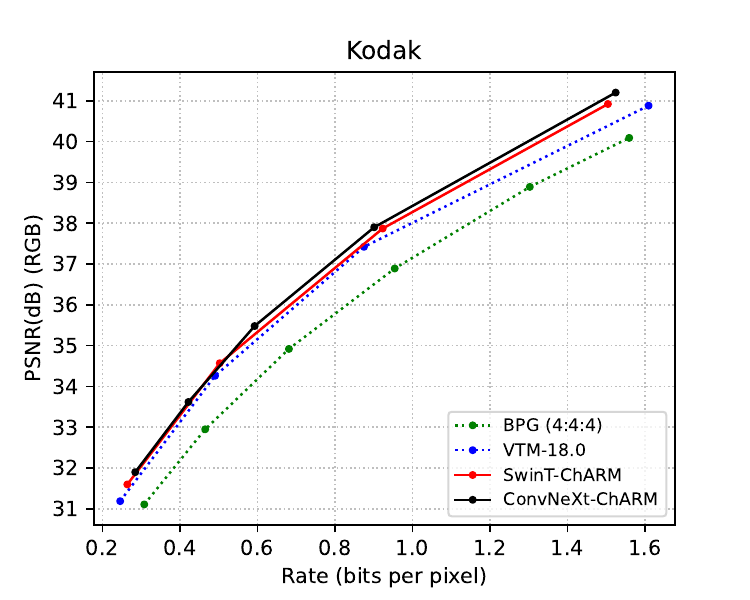}
\caption{Rate-distortion comparison on Kodak dataset.}
\label{rdcurve_kodak}
\end{figure}

To demonstrate the compression efficiency of our proposed approach, we visualize the rate-distortion curves of our model and the baselines on each of the considered datasets.
Considering the Kodak dataset, Figure~\ref{rdcurve_kodak} shows that our ConvNeXt-ChARM outperforms the state-of-the-art learned approach SwinT-ChARM, as well as the \ac{bpg}(4:4:4) and VTM-18.0 traditional codecs in terms of PSNR. Regarding rate savings over VTM-18.0, SwinT-ChARM has more compression abilities only for low PSNR values.
Our model can be generalized to high resolution image datasets (Tecnick, JPEG-AI, and CLIC21), and can still outperform existing traditional and the learned image compression method SwinT-ChARM in terms of PSNR.
Besides the rate-distortion curves, we also evaluate different models using Bjontegaard's metric \cite{bjontegaard2001calculation}, which computes the average bitrate savings (\%) between two rate-distortion curves. 
In Table~\ref{bdrate}, we summarize the BD-rate of image codecs across all four datasets compared to the VTM-18.0 as the anchor.
On average, ConvNeXt-ChARM is able to achieve 5.24\% rate reduction compared to VTM-18.0 and 1.22\% relative gain from SwinT-ChARM.
Figure~\ref{bdr_vs_dt} shows the BD-rate (with VTM-18.0 as an anchor) versus the decoding time of various approaches on the Kodak dataset. It can be seen from the figure that our ConvNeXt-ChARM achieves a good tradeoff between BD-rate performance and decoding time.
\begin{table}[!htb]
\centering
\small
\caption{BD-rate$\downarrow$ performance of \ac{bpg} (4:4:4), SwinT-ChARM, and ConvNeXt-ChARM compared to the VTM-18.0 for the four considered datasets.}\label{bdrate}
\begin{tabular}{@{}l|ccc@{}}
\toprule
\textbf{Dataset}  & \textbf{BPG444} & \textbf{SwinT-ChARM} & \textbf{ConvNeXt-ChARM} \\
\midrule
Kodak             & 20.73\% & -3.47\% & \textbf{-4.90\%} \\
Tecnick           & 27.03\% & -6.52\% & \textbf{-7.56\%} \\
JPEG-AI           & 28.14\% & -0.23\% & \textbf{-1.17\%} \\
CLIC21            & 26.54\% & -5.86\% & \textbf{-7.36\%} \\
\midrule
Average           & 25.61\% & -4.02\% & \textbf{-5.24\%} \\
\bottomrule
\end{tabular}%
\vspace{-4mm}
\end{table}

\begin{table}[!htbp]
\centering
\caption{Image codec complexity. We calculated the average decoding time across 7000 images at 256$\times$256 resolution, encoded at 0.6 bpp. The best score is highlighted in bold.}\label{complexity}
\adjustbox{max width=0.98\textwidth}{%
\begin{tabular}{@{}l|cc|c|c@{}}
\toprule
\multirow{2}{*}{Image Codec} & \multicolumn{2}{c|}{{Latency(ms)$\downarrow$}} & \multirow{2}{*}{GFLOPs$\downarrow$} & \multirow{2}{*}{\#params(M)$\downarrow$}   \\ \cmidrule{2-3}
& GPU  & CPU & &  \\
\midrule
Conv-ChARM            & 124.32          & 967.43           & {\bf 117} & 123.84 \\
SwinT-ChARM           & \textbf{102.45} & 1088.16          & 122       & 127.78 \\
Ours & 122.70          & \textbf{834.42}  & 119       & \textbf{122.33} \\
\bottomrule
\end{tabular}%
\vspace{-4mm}
}\end{table}
\subsection{Models Scaling Study}
\begin{figure}[t]
\centering
\includegraphics[width=0.5\textwidth]{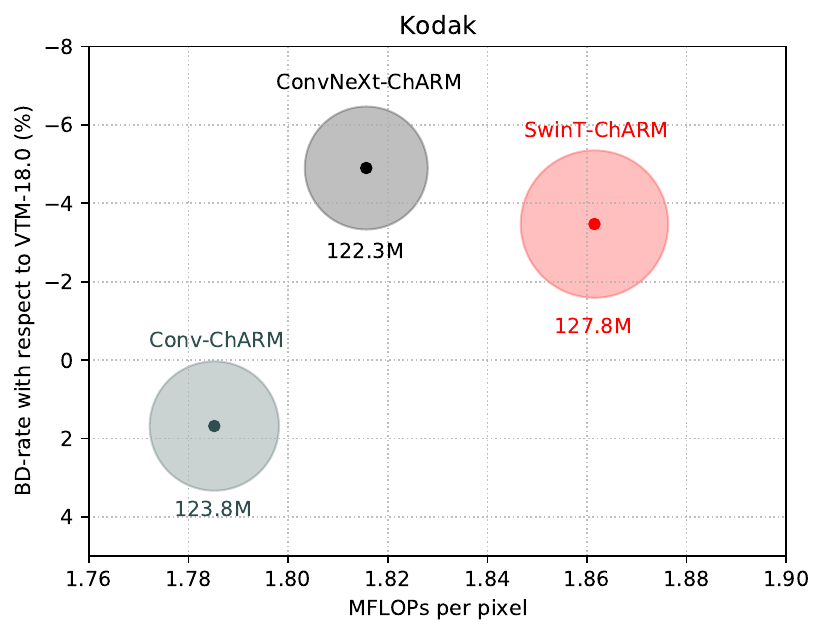}
\caption{Model size scaling. BD-Rate versus MFLOPs per pixel for our model ConvNeXt-ChARM compared to Conv-ChARM and SwinT-ChARM (for both encoding and decoding).}
\label{bdr_vs_flops}
\end{figure}

We evaluated the decoding complexity of the three considered image codecs by averaging decoding time across 7000 images at 256$\times$256 resolution, encoded at 0.6 bpp. We present the image codec complexity in Table~\ref{complexity}, including decoding time on GPU and CPU, floating point operations per second (GFLOPs), the memory required by model weights, and the total model parameters. The models run with Tensorflow 2.8 on a workstation with one RTX 5000 Ti GPU. The Conv-ChARM model refers to the Minnen \etal \cite{minnen2020channel} architecture with a latent depth of 320 and a hyperprior depth of 192, and can be considered as ablation of our model without ConvNeXt blocks. We maintained the same slice transform configuration of the \ac{charm} for the three considered models. The total decoding time of SwinT-ChARM decoder is less than \acp{convnet}-based decoder on GPU but is the highest on CPU. Our ConvNeXt-ChARM is lighter than the Conv-ChARM in terms of the number of parameters, which proves the ConvNeXt block's well-engineered design. Compared with SwinT-ChARM, our ConvNeXt-ChARM shows lower complexity, requiring lower training time with less memory consumption. In addition, Figure~\ref{bdr_vs_flops} shows that our method is in an interesting area, achieving a good tradeoff between BD-rate score on Kodak, total model parameters, and MFLOPs per pixel, highlighting an efficient and hardware-friendly compression model. 

\subsection{Comparison with SwinT-ChARM}
ConvNeXt-ChARM achieves good rate-distortion performance while significantly reducing the latency, which is potentially helpful to conduct, with further optimizations, high-quality real-time visual data transmission, as recently proposed in the first software-based neural video decoder running HD resolution video in real-time on a commercial smartphone \cite{le2022mobilecodec}.
Since fewer works attempt to explicitly compare Swin Transformer and \ac{convnet}-based blocks, here, we compare our ConvNeXt-ChARM with SwinT-ChARM under the same conditions and configurations. We found that a well-designed \ac{convnet}, without any additional attention modules, can outperform the highly coveted Swin Transformer in learned transform coding in terms of BD-rate, with more visually pleasing reconstructions and comparable decoding latency. In addition, ConvNeXt-ChARM maintains the efficiency and maturity of standard \acp{convnet} and the fully-convolutional nature for both training and inference.
There is no doubt that Transformers are excellent architectures with enormous potential for the future of various computer vision applications. However, their vast hunger for data and computational resources~\cite{zhu2020deformable} poses a big challenge for the computer vision community. Taking SwinT-ChARM as an example, it needs, on average, $\times$1.33 more time than ConvNeXt-ChARM, to train on the same number of epochs.

\section{Conclusion}
\label{concl}
In this work, we reconcile compression efficiency with ConvNeXt-based transform coding paired with a \ac{charm} prior and propose an up-and-coming learned image compression model ConvNeXt-ChARM. Furthermore, we inherit the advantages of pure \acp{convnet} in the proposed method to improve both efficiency and effectiveness. The experimental results, conducted on four datasets, showed that our approach outperforms previously learned and conventional image compression methods, creating a new state-of-the-art rate-distortion performance with a significant decoding runtime decrease.
Future work will further investigate efficient low-complexity entropy coding approaches to further enhance decoding latency. 
With the development of GPU chip technology and the further optimization of engineering, learning-based codecs will be the future of coding, achieving better compression efficiency when compared with traditional codecs and aiming to bridge the gap to a real-time operation.
We hope our study will challenge certain accepted notions and prompt people to reconsider the significance of convolutions in computer vision. 

\bibliographystyle{IEEEbib}
\bibliography{refs}

\end{document}